%% file: root.tex
\documentclass[letterpaper, 10 pt, conference]{ieeeconf}  % Comment this line out if you need a4paper

\IEEEoverridecommandlockouts                              % This command is only needed if 
\usepackage[T1]{fontenc}
\usepackage{color}
\usepackage{algorithm,algcompatible}

\usepackage{algpseudocode}
\usepackage{cite}
\usepackage{hyperref}

\input{boldfacemath.tex}

\usepackage{graphicx}

\def\VelocityTracking{\centering{
\includegraphics[width=0.45\textwidth]{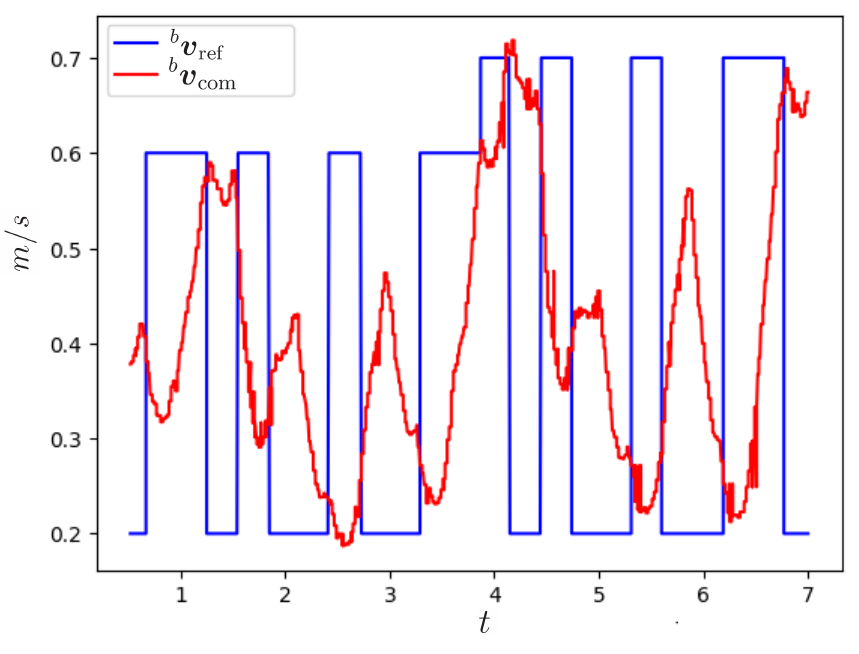}}}

\def\BlockScheme{\centering{
\includegraphics[width=0.95\textwidth]{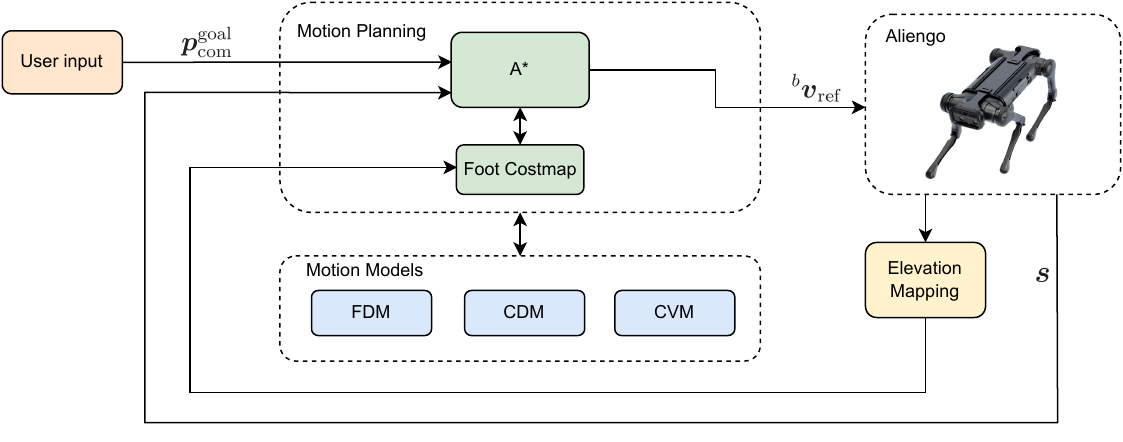}}}

\def\SnapshotsTop{\centering{
\includegraphics[width=0.22\textwidth]{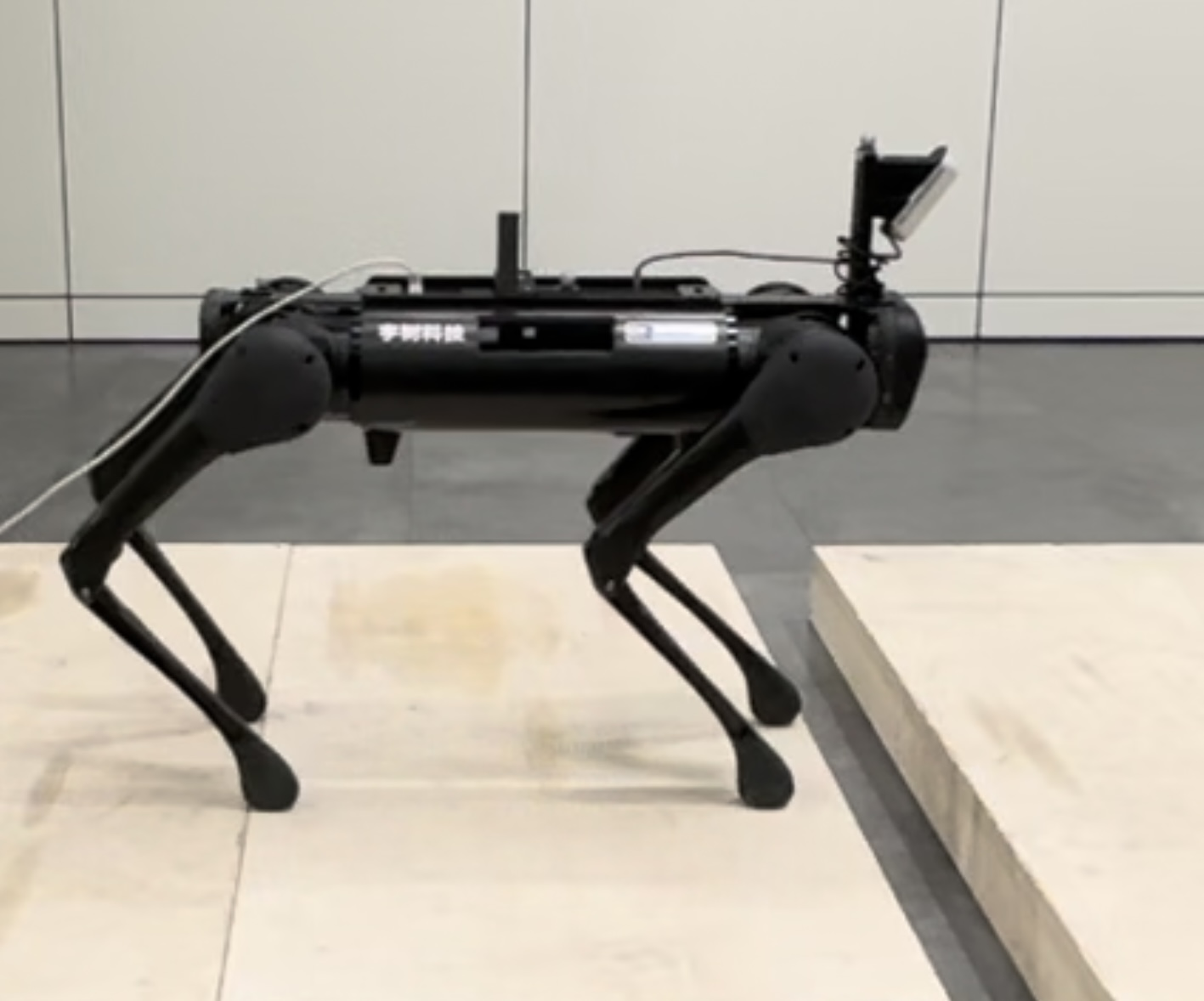}
\includegraphics[width=0.22\textwidth]{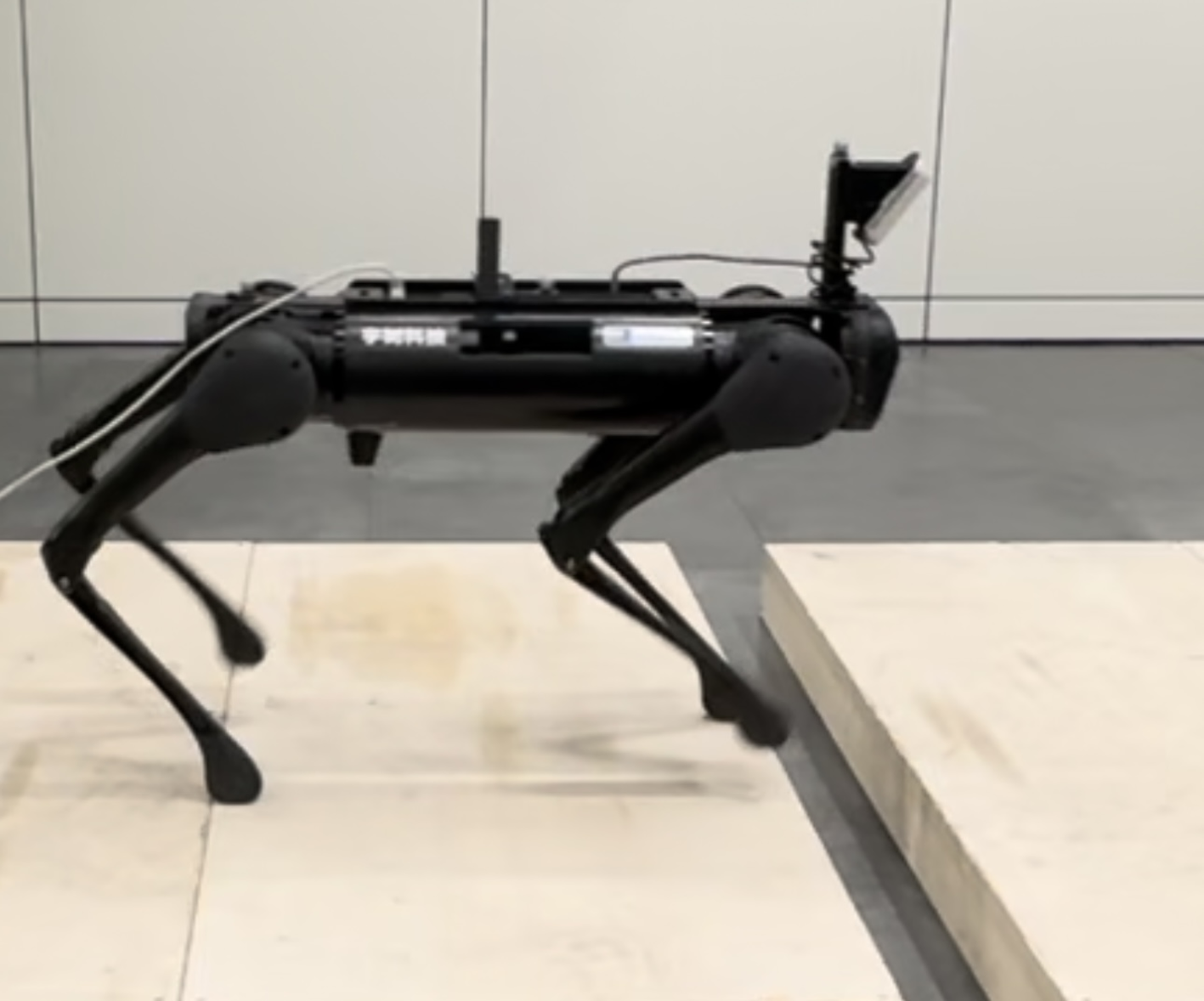}}}

\def\SnapshotsBottom{\centering{
\includegraphics[width=0.22\textwidth]{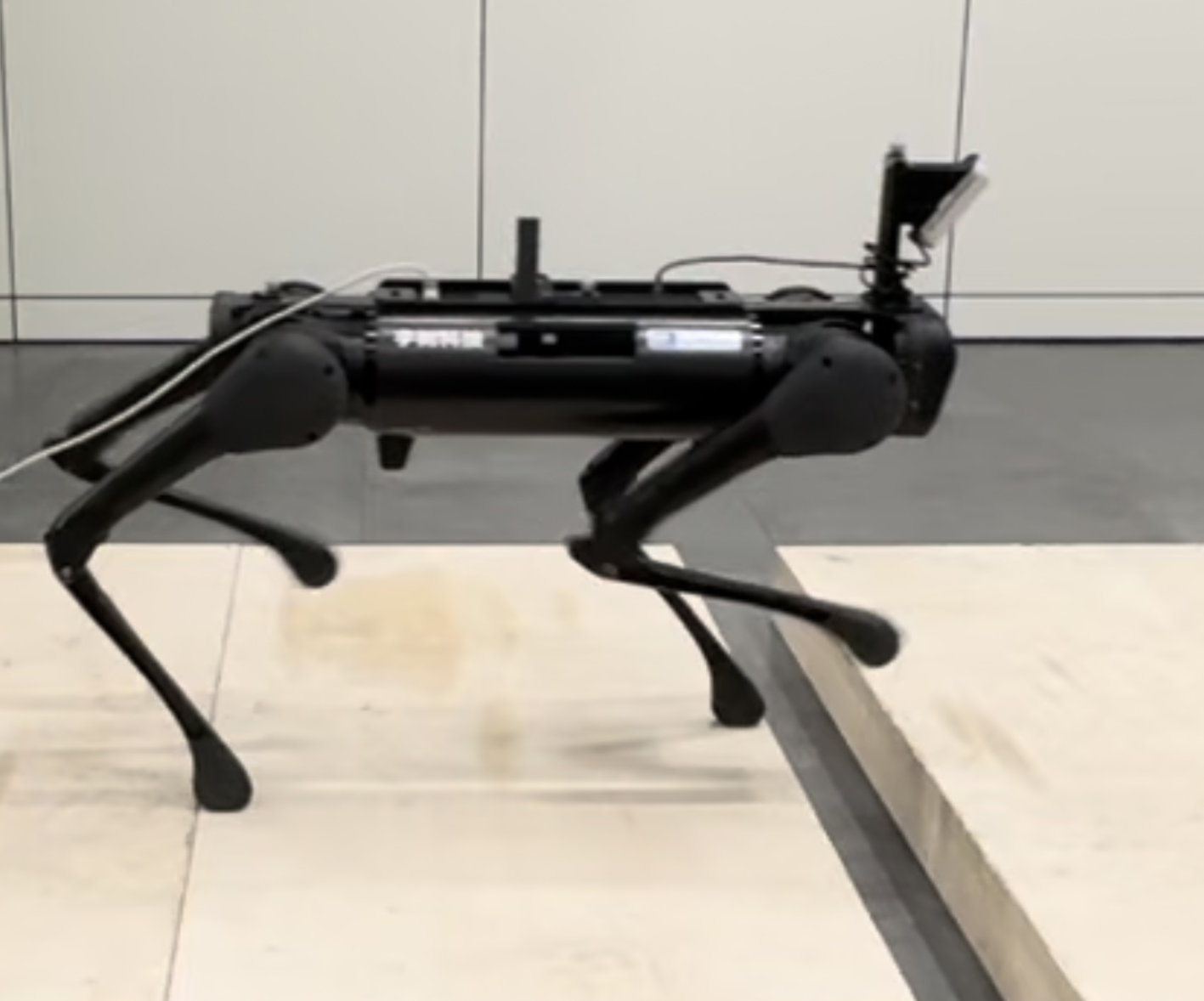}
\includegraphics[width=0.22\textwidth]{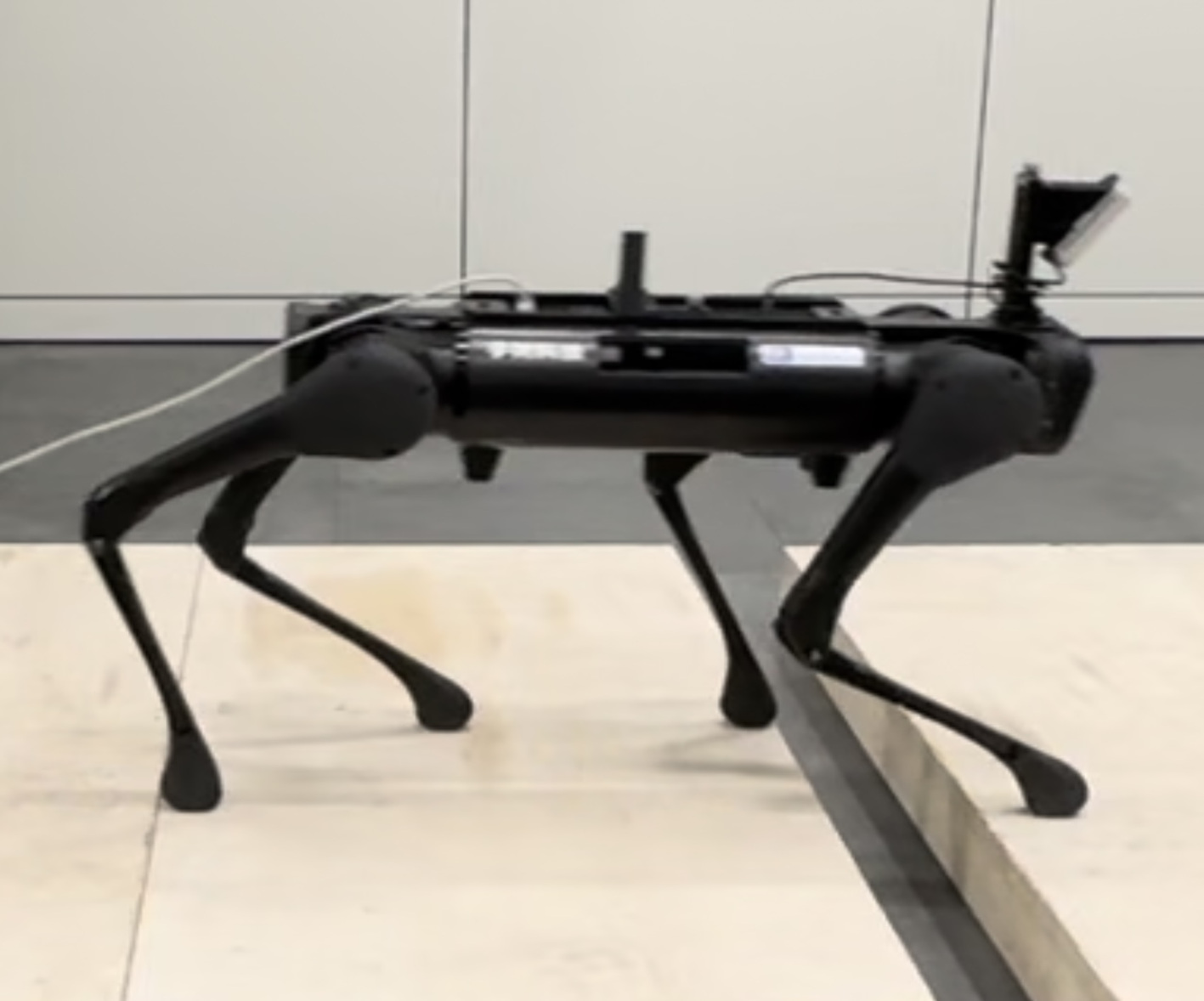}}}

\def\Scenarios{\centering{
\includegraphics[width=0.32\textwidth]{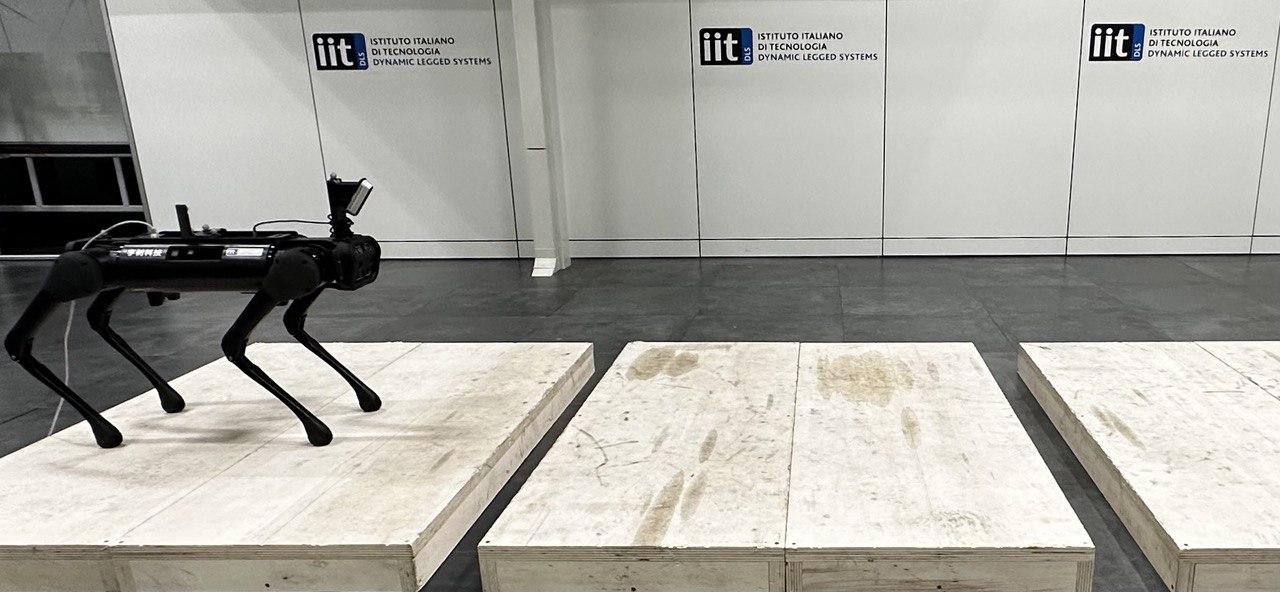}
\includegraphics[width=0.32\textwidth]{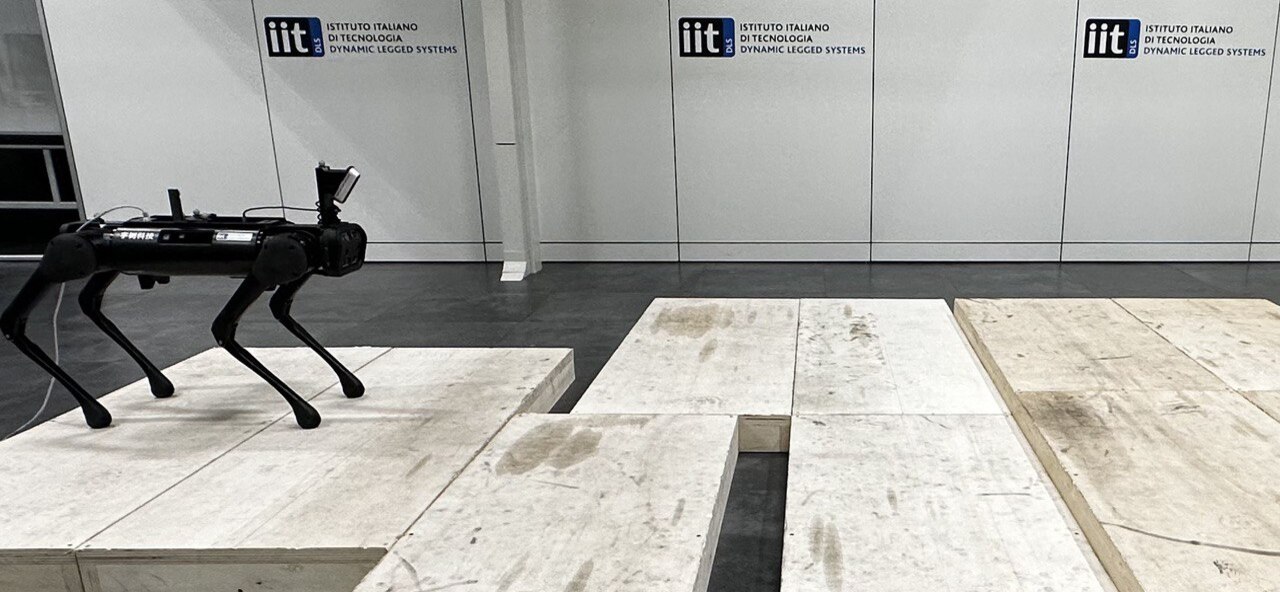}
\includegraphics[width=0.32\textwidth]{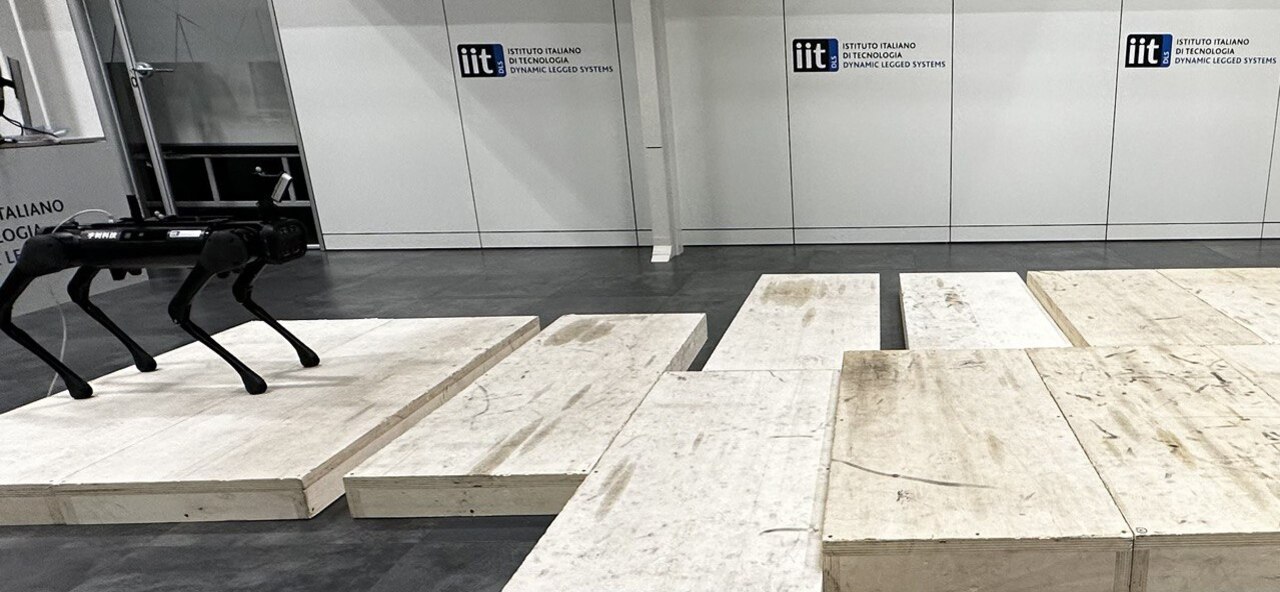}}}

\def\Costmaps{\centering{
\includegraphics[width=0.32\textwidth]{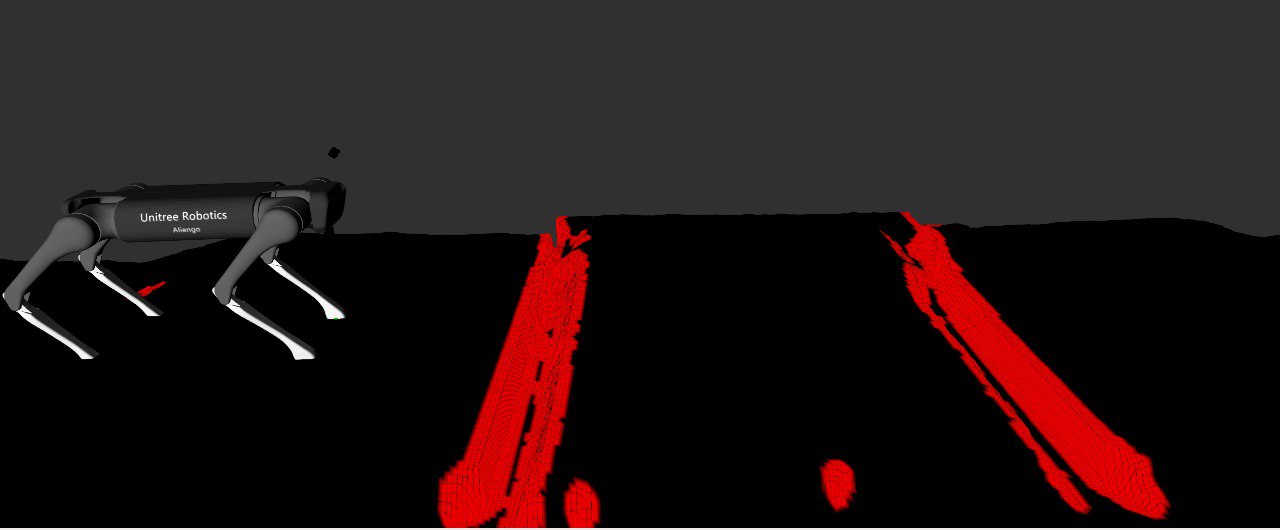}
\includegraphics[width=0.32\textwidth]{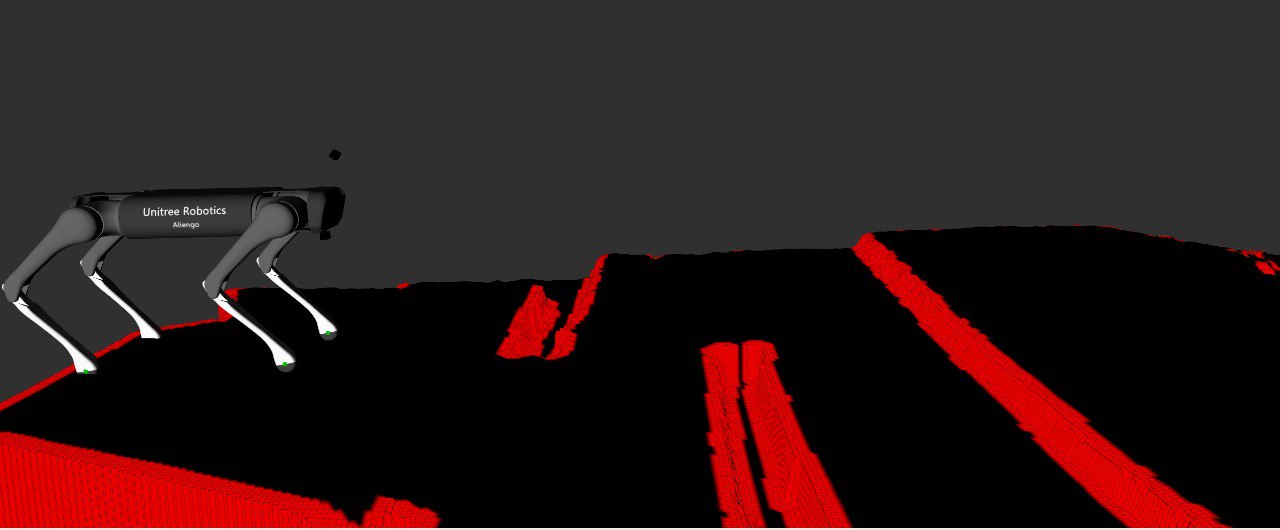}
\includegraphics[width=0.32\textwidth]{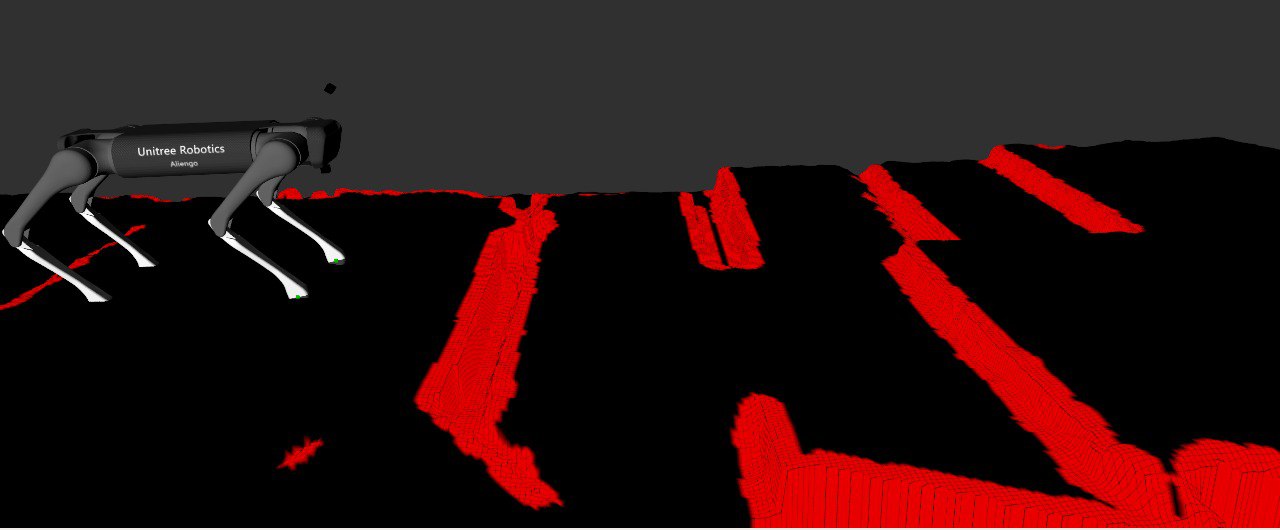}}}

\usepackage{xcolor}
\usepackage{eso-pic}
\AddToShipoutPictureBG*{%
  \AtPageUpperLeft{%
    \setlength{\unitlength}{1mm}%
    \put(-9.5,-9){\makebox(\paperwidth,0)[c]{\parbox{0.9\textwidth}{2023 IEEE/RSJ International Conference on Intelligent Robots and Systems (IROS), pp. 800-806}}}
  }
}

\AddToShipoutPictureBG*{%
  \AtPageUpperLeft{%
    \setlength{\unitlength}{1mm}%
    \put(-9.5,-14){\makebox(\paperwidth,0)[c]{\parbox{0.9\textwidth}{DOI: \href{https://ieeexplore.ieee.org/document/10342440}{10.1109/IROS55552.2023.10342440}}}}
  }
}

\title{\LARGE \bf
Quadrupedal Footstep Planning using Learned
Motion Models of a Black-Box Controller
}

\author{Ilyass Taouil$^{1,2}$, Giulio Turrisi$^{1}$, Daniel Schleich$^{2}$, Victor Barasuol$^{1}$, Claudio Semini$^{1}$, and Sven Behnke$^{2}$% <-this % stops a space
\thanks{$^{1}$Dynamic Legged Systems Laboratory, Istituto Italiano di Tecnologia (IIT), 16163 Genova, Italy}%
\thanks{$^{2}$Autonomous Intelligent Systems, University of Bonn, 53115 Bonn, Germany}%
}

\begin{document}

\maketitle
\thispagestyle{empty}
\pagestyle{empty}

%%%%%%%%%%%%%%%%%%%%%%%%%%%%%%%%%%%%%%%%%%%%%%%%%%%%%%%%%%%%%%%%%%%%%%%%%%%%%%%%
\begin{abstract}

Legged robots are increasingly entering new domains and applications, including search and rescue, inspection, and logistics. However, for such a systems to be valuable in real-world scenarios, they must be able to autonomously and robustly navigate irregular terrains. In many cases, robots that are sold on the market do not provide such abilities, being able to perform only blind locomotion. Furthermore, their controller cannot be easily modified by the end-user, requiring a new and time-consuming control synthesis. In this work, we present a fast local motion planning pipeline that extends the capabilities of a black-box walking controller that is only able to track high-level reference velocities. More precisely, we learn a set of motion models for such a controller that maps high-level velocity commands to Center of Mass (CoM) and footstep motions. We then integrate these models with a variant of the  $A^*$ algorithm to plan the CoM trajectory, footstep sequences, and corresponding high-level velocity commands based on visual information, allowing the quadruped to safely traverse irregular terrains at demand.

\end{abstract}

%%%%%%%%%%%%%%%%%%%%%%%%%%%%%%%%%%%%%%%%%%%%%%%%%%%%%%%%%%%%%%%%%%%%%%%%%%%%%%%%
\section{Introduction}

\begin{figure}[!ht]
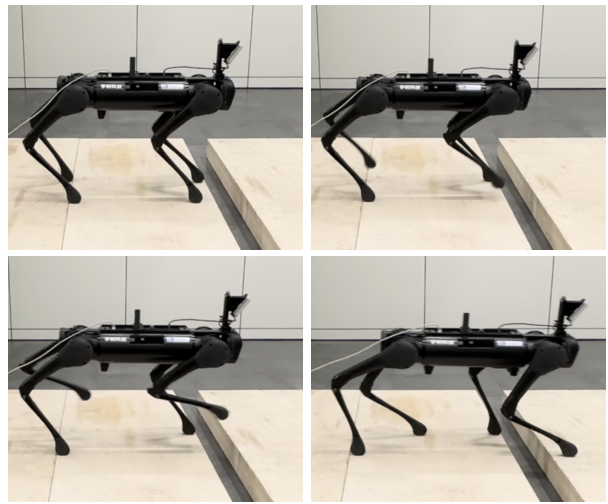

\SnapshotsTop
\vspace{1mm}
\SnapshotsBottom
\caption{The Aliengo quadruped robot stepping over a gap with the proposed approach.}
\vspace{-5mm}
\label{figure:snapshots}
\end{figure}

Autonomous systems are continuously entering new domains and applications. Significant leaps have been made in making these systems operational, demonstrating enormous potential for the future. Legged systems in particular offer high mobility and versatility given their ability to step at discontinuous locations. On the other hand, to be able to be deployed in the real world they must be able to reason about the surroundings in order to select the best contact locations. The work we present addresses the development of perceptive locomotion skills for legged robots whose manufacturer's controller can only track high-level velocity commands using proprioceptive data, and cannot receive target footstep locations. Examples of such a systems are Aliengo~\cite{aliengo}, Vision 60~\cite{ghost}, and Spot~\cite{spot}. We seek to extend the capabilities of such controllers by planning the appropriate velocity commands that allow the legged system to overcome different types of uneven terrains (Fig.~\ref{figure:snapshots}). We do so by using exteroceptive sensory data and a set of motion models learned from the controller's behavior, of which we do not know the implementation details, hence treating it as a black box.

In the literature, there exists several paradigms to tackle vision-based locomotion. In \cite{belter2016adaptive, mastalli2020motion, fankhauser2018robust, griffin2019footstep, villarreal2020mpc, kim2020vision, grandia2022perceptive} the authors use a model-based approach to plan the required footstep sequences to overcome irregular terrains. Learning-based methods \cite{gangapurwala2022rloc, peng2017deeploco, tsounis2020deepgait} have also shown great ability to couple control and perception in these scenarios. However, the aforementioned approaches, respectively require the ability to either be able to set the desired footstep location at the controller level, or to have available in simulation the behavior of the closed-loop system to learn the policy, something that in both cases is not available on the previously mentioned commercial platforms.

Works similar to ours are the one presented in~\cite{chestnutt2005footstep,schmitz2012real,stumpf2014supervised}. In~\cite{chestnutt2005footstep}, the authors presented a footstep planner for the humanoid robot ASIMO~\cite{sakagami2002intelligent} whose controller can only receive desired body displacements. The authors use the $A^*$~\cite{hart1968formal} search algorithm to compute the optimal sequence of footstep locations up to a pre-defined planning horizon, using a mapping in the form of a lookup table where footstep displacements are defined based on the current state, action, and environment. The method, however, is limited to slow and static gaits given the high re-planning time. Furthermore, it can only cope with flat terrains. In~\cite{schmitz2012real}, the authors presented a method that generates footstep trajectories for a humanoid to approach and kick a ball. Similar to the previous method, the authors only consider flat terrains and their $A^*$ based planner requires several minutes to find an optimal solution, hence requiring a subsequent approximation of the final solution by policy learning. Finally, in~\cite{stumpf2014supervised}, the authors presented an approach that plans a sequence of footsteps for the Atlas humanoid, which is commanded by a black-box controller. However, their method allows changing the footstep locations at the controller level, which in our study case is not possible.

To summarize, the main contributions of this paper are the following:
\begin{itemize}
  \item a novel motion planning pipeline for quadrupedal locomotion that extends blind locomotion capabilities of black-box controllers with perception-based skills
  \item a set of learned motion models that predict the behavior of a black-box quadrupedal locomotion controller
  \item extensive experimental tests with the Aliengo robot over different test scenarios that demonstrate the capability of the approach
\end{itemize}

\subsection{Outline}
\label{subsec:outline}

The paper is organized as follows: Section~\ref{sec:formulation} describes the problem formulation, whereas Section~\ref{sec:approach} delves into the details of the proposed method and describes the learned motion models and the footstep planning formulation. Section~\ref{sec:results} introduces the metrics used for the evaluation and presents the results obtained from the extensive experimentation in different indoor scenarios. Finally, Section~\ref{sec:conclusions} summarizes the presented method and offers some considerations on its limitations and possible future extensions.

\section{Problem Formulation}
\label{sec:formulation}

\begin{figure*}[!t]
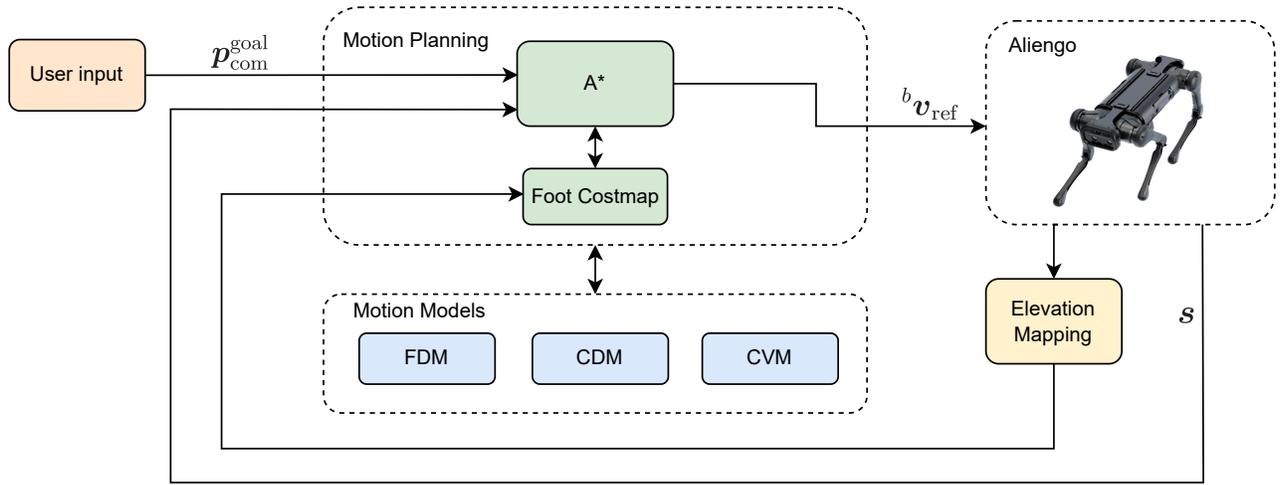

\BlockScheme
\caption{Block diagram of the overall planning procedure. Every time a full-stance phase is detected, a new solution is computed in a receding horizon fashion. For clarity, we dropped the dependence on the time $t$. The motion models are:
Footstep Displacement Model (FDM), CoM Displacement Model (CDM), and CoM Velocity Model (CVM).}
\label{fig:overview_method}
\vspace{-10pt}
\end{figure*}

In this work, we address the development of perceptive locomotion skills for purely reactive controllers. Such controllers solely rely on proprioceptive information in order to keep balance, thereby making them prone to failure in complex scenarios where vision is essential. The Aliengo robot, which will be under study in this work, is an example of such a system, as its built-in controller can only perform blind locomotion and can only be commanded with high-level velocity references. 

Given the limited control authority that the user can have on such a system, the proposed solution is formulated as a motion planning problem that aims at finding the optimal footstep sequences by varying velocity commands. In fact, CoM velocities and footstep positions are quantities often correlated~\cite{first_foot, second_foot, raibert1986legged}, meaning that this mapping, if learned, can be employed to allow Aliengo to traverse irregular terrain at demand.

The API of Aliengo provides multiple information about the robot states, such as the robot CoM position and velocity, defined as $\pv_{\textrm{com}}$ and ${}^b\vv_{\textrm{com}}$ and respectively defined in the world frame $W$ and in the base frame $B$, the relative position of the feet $\bar{\pv}_{\textrm{foot}}$ with respect to the CoM, the robot contact forces $\fv$, and the rotation matrix that maps information from the robot base to the world frame ${}^w\Rm_b$. All these quantities, as will be explained in Section~\ref{sec:approach}, will be exploited to learn accurate motion models of the locomotion controller of Aliengo. We also have access to depth information from a top mounted RealSense D435f camera to reconstruct the environment.

Furthermore, the blind controller under study only allows a trotting gait, where two diagonal feet swing at a time. This will inevitably hinder the possible scenarios that can be traversed, constraining the two swing legs to have the same foot displacement at the same time. This limitation can only be alleviated by a planning procedure with a sufficient horizon length.

\section{Proposed Approach}
\label{sec:approach}

The motion planning pipeline consists of three blocks: 1) a motion models interface that uses the learned models to predict the robot's next CoM position, CoM velocity, and swing leg touchdown; 2) an elevation mapping module~\cite{Fankhauser2018ProbabilisticTerrainMapping} that reconstructs the local terrain as a $2.5D$ grid map, which is then used to compute a foot costmap describing safe and unsafe stepping locations; 3) a motion planning module that implements a variant of the $A^*$ algorithm and integrates the output of the previous two modules to plan footstep sequences and corresponding high-level velocities, at every detected full-stance phase, which are then fed to the robot controller.

At a high level, the above motion planning pipeline works
as follows: the algorithm, given a user-defined CoM goal position $\pv^{\textrm{goal}}_{\textrm{com}}$, acquires the actual robot state at the current time $t$, defined as 
\begin{equation}
\sv^{t} = (\pv^t_{\textrm{com}}, {}^b\vv^t_{\textrm{com}}, {}^b\vv^{t-1}_{\textrm{com}},  {}^b\vv^t_{\textrm{ref}}, {}^b\vv^{t-1}_{\textrm{ref}},    {}^b\bar{\pv}^t_{\textrm{foot}}) 
\label{eq:state}
\end{equation}
where ${}^b\vv^t_{\textrm{ref}}$, ${}^b\vv^{t-1}_{\textrm{ref}}$ are the last two commanded velocities, and ${}^b\vv^{t-1}_{\textrm{com}}$ is the previous CoM velocity. The planner then iterates over a set of discrete velocity references and starts building a graph describing the evolution of the robot states until $\pv^{\textrm{goal}}_{\textrm{com}}$ is reached or $N$ sequential safe footholds are found. For this, three different regressor models are used to predict the components of $\sv$ during each node expansion: a CoM Displacement Model (CDM), a Footstep Displacement Model (FDM), and a CoM Velocity Model (CVM). 
Finally, once the solution is obtained, only the first velocity command is applied to the robot and then the procedure is restarted in a receding horizon fashion.

A block diagram of the proposed approach can be seen in Fig.~\ref{fig:overview_method}.

In the following sections, we describe the learning procedure to obtain the models (Section~\ref{subsec:models}), detailing the necessity of the extended state representation adopted in~(\ref{eq:state}), and the proposed planning algorithm (Section~\ref{subsec:planner}).

\subsection{Motion Models Learning}
\label{subsec:models}

The motion models learned in this work map an input vector describing a reduced representation $\xv^t$ of the robot state in~(\ref{eq:state}), to feet and CoM displacement as well as the CoM velocity at time $t+1$, fundamentally creating three different regression problems. Different approaches could be used to learn these models, including a simple \emph{mapping table} or a \emph{Neural Network}. However, given that in our experiments we observed a strong linear character exhibited by the controller (see Sect.~\ref{sec:results}), a better approach is found in the \emph{Multivariate Linear Regression} (MLR) \cite{su2012linear}, where a linear combination in the form of a first-degree polynomial that fits the data is computed.

%\gtur{What follows is a characterization of such models and an explanation of the data acquisition and training process.}
The most fundamental variable we want to predict in our planning pipeline is the next footstep placement $\pv^{t+1}_{\textrm{foot}}$. In our application, the FDM regressor outputs a displacement ${}^b\deltav \pv_{\text{foot}}^t$ over the actual foot position expressed in the base frame. Usually, this variable is strictly correlated with the commanded and actual CoM velocity of the robot, but other variables can also play an important role in its derivation. For this reason, we experimented with different input spaces for this model, and the best accuracy was obtained by choosing it as
\begin{equation}
\xv^t_{\textrm{FDM}} = ({}^b\vv^t_{\textrm{com}}, {}^b\vv^{t-1}_{\textrm{com}},  {}^b\vv^t_{\text{ref}}, {}^b\vv^{t-1}_{\textrm{ref}}, {}^b\bar{\pv}^t_{\textrm{foot}}). 
\label{eq:state2}
\end{equation}
As will be explained in Sect.~\ref{sec:results}, this input space choice is by no means dependent on the type of the regressors, since we observed that employing both linear and nonlinear basis functions does not impact the overall final accuracy. 

\begin{figure}[!b]
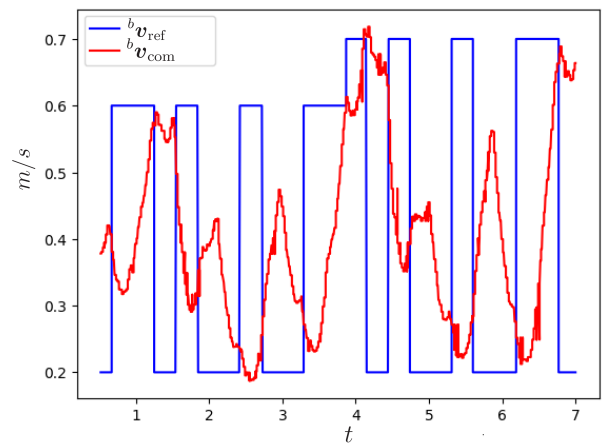

\VelocityTracking
\caption{The CoM velocity reference tracking behavior in $x$ of the Aliengo's built-in controller. In this work, the controller is set in the "Sport Mode 3.0".}
\label{fig:controller_behavior_sp}
\end{figure}

The input space $\xv^t_{\textrm{FDM}}$ is characterized by the presence of the commanded and actual CoM velocities at time $t$ and time $t-1$ as well. We found that this state history information is needed in order to mimic the non-ideal behavior of the controller of Aliengo, which is unable to track the commanded velocity in just one foot swing (Fig.~\ref{fig:controller_behavior_sp}). 

In our approach, we plan with a look-ahead horizon of $N$ footsteps in order to reduce myopic behaviors in the final solution. Thus, we need to predict the future state variables ${}^b\vv^{t+i}_{\textrm{com}}, {}^b\bar{\pv}^{t+i}_{\textrm{foot}}$, with $i = 1, ..., N$, for the FDM state representation~(\ref{eq:state2}) as well. For this, the CVM regressor provides directly the future CoM velocity ${}^b\vv^{t+i}_{\textrm{com}}$, while the CDM regressor predicts the displacement ${}^b\deltav \pv^t_{\text{com}}$ with respect to the actual CoM position. This value, together with the previous output of the FDM regressor, is then used to retrieve the last component of $\xv^{t+1}_{\text{FDM}}$ as the following: first, we compute the next CoM and foot position in the world frame as

\begin{equation}
 \pv_{\textrm{foot}}^{t+1} = {}^w\pv_{\textrm{foot}}^t + {}^w\Rm_b{}^b\deltav \pv_{\text{foot}}^t
 \label{eq:foot_world}
\end{equation}
\begin{equation}
\pv_{\textrm{com}}^{t+1} = \pv_{\textrm{com}}^{t} + {}^w\Rm_b {}^b\deltav \pv_{\text{com}}^{t} 
\label{eq:com_world}
\end{equation}
%where ${}^w\Rm_b$ is the rotation matrix from the base frame to the world frame, and 
and then we evaluate their relative displacement as
\begin{equation}
\begin{aligned}
  {}^b&\bar{\pv}_{\textrm{foot}}^{t+1} = {}^b\Rm_w (\pv_{\textrm{foot}}^{t+1} - \pv_{\textrm{com}}^{t+1})  
\end{aligned}
\label{eq:foot_displacement}
\end{equation}

Both CVM and CDM regressors share the same input space with the FDM.

For training the above models, we first performed a data acquisition step gathering the quadruped's high-level state information offered by the API, while it executes various motion commands. The data collection was performed on a flat ground with a step height of $10$cm. To acquire the data we use Aliengo's controller interface that receives as input a $3$-dimensional velocity command. Based on this interface, we defined a strategy to gather the data such that they reflect all possible behaviors of interest to the planning process. This entails exhausting all possible input permutations that can be commanded to the robot in terms of velocity magnitudes. More precisely, we send several reference commands over varying periods of time to learn the controller's behavior during acceleration, deceleration, and continuous motions.
Finally, we fit the corresponding MLR models by applying a simple least square loss.

\subsection{Footstep Planning with Learned Models}
\label{subsec:planner}
The motion planning process consists of finding the best sequence of high-level velocity commands that can be fed to the quadruped's internal controller in order to overcome complex terrains. To achieve this task, we utilize a variant of the graph-based $A^*$ algorithm~\cite{dolgov2008practical} that is able to better cope with the continuous state-space nature of the footstep planning problem in hand. It should be noticed that other planning methods could have been employed as well, such as randomized algorithms~\cite{RRT}, but graph-based search algorithms usually show a superior converge rate to a solution for tasks with small-dimensional state spaces.   

Starting from a generic full-stance phase defined by the state $\sv^{\textrm{start}}$, where all the feet of the robot are in contact with the ground, the planner searches for the best $N$ future footsteps needed to reach a user-defined goal position $\pv^{\textrm{goal}}_{\textrm{com}}$. For this, first we expand a set of discretized velocities to generate $M$ new nodes $\sv^{_{'}}$, with $M$ the possible velocity command values that we can apply to our robot. In our application, we discretize the action space with a step-size of 0.1 $m/s$, and we set the minimum and maximum allowed command velocity, respectively at 0.0 $m/s$ and 1 $m/s$. Afterwards, we compute the state information associated with all the expanded nodes. This step is needed to check if those are associated with states that were visited before and if the associated footholds are safe.

For this,  we first compute the CoM position $\pv_{\textrm{com}}^{_{'}}$ of every single node in the world frame by applying the CDM regressor and Eq.~(\ref{eq:com_world}).
We then discretize $\pv_{\textrm{com}}^{_{'}}$ with a 0.01 $m$ step-size. To check if the new nodes are redundant, we maintain a list of visited nodes $V$. The expanded nodes are directly discarded if there is a node with the same velocity command and discretized CoM position value, otherwise a safety check is performed on the proposed foothold locations. For this, we compute the associated footstep displacements ${}^b\deltav \pv_{\text{foot}}$ via the FDM regressor and compute the respective world coordinate $\pv_{\textrm{foot}}^{_{'}}$ by Eq.~(\ref{eq:foot_world}).

 \begin{algorithm}
     \caption{Footstep planner at the generic time $t$}\label{alg:pseudo}
         \begin{algorithmic}[1]
         \Procedure{Plan}{$\pv^t_{\textrm{com}}, {}^b\vv^t_{\textrm{com}}, {}^b\vv^{t-1}_{\textrm{com}},  {}^b\vv^t_{\textrm{ref}}, {}^b\vv^{t-1}_{\textrm{ref}},    {}^b\bar{\pv}^t_{\textrm{foot}}$}
            
             \State $i \gets {0}$
             %\State ${}^b\vv^{t-1}_{\textrm{com}}, {}^b\vv^{t-1}_{\textrm{ref}}  \gets {0}$
             \State $s^{start} \gets (\pv^t_{\textrm{com}}, {}^b\vv^t_{\textrm{com}}, {}^b\vv^{t-1}_{\textrm{com}},  {}^b\vv^t_{\textrm{ref}}, {}^b\vv^{t-1}_{\textrm{ref}},    {}^b\bar{\pv}^t_{\textrm{foot}}, i)$
             \State $O \gets \{s^{start}\}$
             \State $V \gets \{\}$
             \State $path \gets \{\}$
             \State $s \gets \textbf{visit least-cost state representation from O}$
            
             \If {$n == s^{goal}$ or $i == N$ or $O == \emptyset$} 
                \State \Return $path \gets \textbf{best sequence of velocities}$
             \EndIf
            
             \For{each velocity}
                 
                 %\State {Predict CoM for state n}
                 %\If {invalid CoM} \State \textbf{goto} {9}.
                 %\EndIf
                
                 \State compute Footstep position by applying Eq.~(\ref{eq:foot_world})
                 \If {invalid feet locations} 
                    \State \textbf{goto} 11.
                 \EndIf
                 \State compute CoM position by applying Eq.~(\ref{eq:com_world})
                 \State compute CoM velocity ${}^b\vv_{\textrm{com}}^{_{'}}$
                
                 \State update number of node parents $i^{_{'}} = i + 1$                 
                 \State construct state transition representation $s^{_{'}}$
                 \State compute total cost $d_{goal} + d_{foot}$ for $s^{_{'}}$

                 \If {$s^{_{'}}$ is a duplicate in $V$} 
                     \State discard $s^{_{'}}$
                 \ElsIf{$s^{_{'}}$ is a duplicate in $O$} 
                     \If {$s^{_{'}}$ has lower cost} 
                     \State $O \gets O \cup {s^{_{'}}}$
                     \EndIf
                 \Else
                     \State $O \gets O \cup {s^{_{'}}}$
                 \EndIf
                 
             \EndFor
            
            \State $V \gets V \cup {s}$
            \State \textbf{goto} 7.
        \EndProcedure
     \end{algorithmic}
 \end{algorithm}

The safety of the footholds is then retrieved by analyzing the value of $\pv_{\textrm{foot}}^{_{'}}$ in the foot costmap, where pixels that exhibit a large height variation compared to their neighborhood are flagged as unsafe. We then calculate the associated traversability costs for each new node to choose the next candidate to visit. This comprises a Euclidean distance cost $d_{\textrm{goal}}$ of the robot CoM position with respect to the goal, and a feet configuration cost $d_{\textrm{foot}}$ needed to maximize the feet distances to the closest unsafe cell, such as
\begin{equation*}
d_{\text {foot}}= \begin{cases}d_{\max}-d_{\text {foot}}, & \text { if } d_{\text {foot }}<d_{\max} \\ 0, & \text { otherwise }\end{cases}
\end{equation*}
where $d_{\max}$ is a user-defined safety margin threshold. In fact, solely relying on the Euclidean cost and the footstep validity check might cause the robot to step too close to the edges due to prediction errors or motion inaccuracies. However, by maximizing the feet distances, we can mitigate such errors. We add the nodes in the open list $O$ for further expansion if there are no nodes with the same CoM discretized positions and with lower cost, otherwise we maintain in $O$ only the best candidates. Finally, we retrieve the last components of the nodes state ${}^b\vv^{_{'}}_{\textrm{com}}, {}^b\bar{\pv}_{\textrm{foot}}^{_{'}}$ by respectively querying the CVM regressor and computing Eq.~(\ref{eq:foot_displacement}).

We can then proceed to extract from $O$ the next node to visit by choosing the one with the lowest cost.

The above planning procedure continues until the goal is reached, $N$ safe footsteps are found, or the maximum computational time is reached. In the last two cases, only the first planned velocity is commanded. The planning procedure is then repeated in a receding horizon fashion as soon as another full-stance phase is detected.

The pseudo-code of the proposed footstep planning algorithm can be found in Algorithm~\ref{alg:pseudo}.

\section{Results}
\label{sec:results}

\begin{figure*}[!ht]
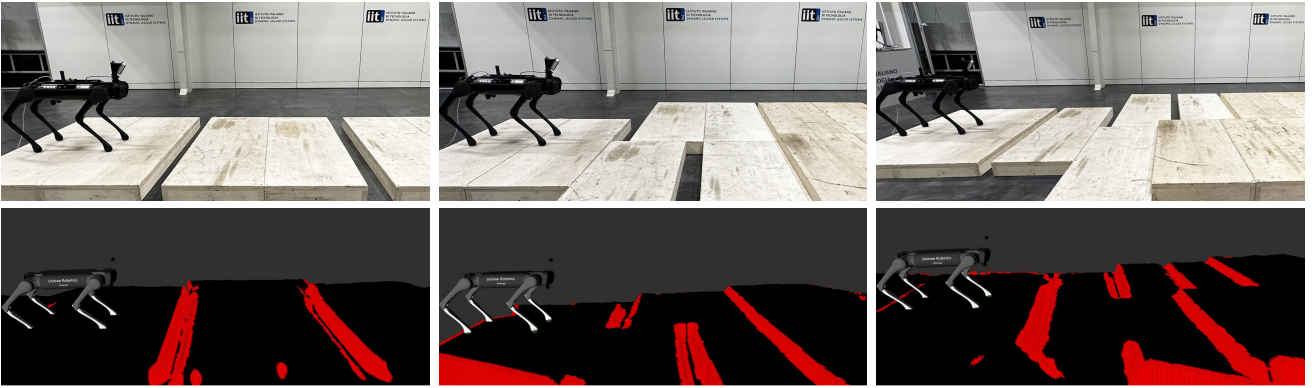

\Scenarios
\vspace{1mm}
\Costmaps
\caption{The proposed test scenarios (top) with the related foot costmaps (bottom). Starting from the left, we show Scenario I, Scenario II, and Scenario III. Furthermore, in the foot costmaps the areas which are not considered safe are shown in red and the traversable areas are shown in black. }
\vspace{-2mm}
\label{figure:scenarios}
\end{figure*}

In this section we present the results of the prediction models and of the motion planning pipeline explained in Section~\ref{sec:approach}. All the experiments are executed on a system equipped with a top-mounted Intel RealSense D435f camera (Fig.~\ref{figure:snapshots}) and an external motion capture system to obtain odometry information and facilitate the computation of the elevation map from which we then compute the related foot costmap.

First, we present an empirical evaluation of the prediction models and a
comparison between the different design choices of the regressors' input space. Afterward, we evaluate the proposed motion planning pipeline using the Aliengo system on three different indoor scenarios, where we neglect rotational movements.

\begin{table}[b]
\vspace{-2mm}
\caption{Regressor models comparison.}
\centering
\begin{tabular}{ |p{5cm}|p{1cm}|}
\hline
\multicolumn{1}{|c|}{\textbf{Regression Models}}&
\multicolumn{1}{c|}{\textbf{RMSE (m)}} \\
\hline
CDM w/o ${}^b\vv^{t-1}_{\text{ref}}, {}^b\vv^{t-1}_{\text{com}}$ & 0.0193 \\
\vspace{-1.8mm}
CDM w/o $\pv_{\text{foot}}^t$ & 0.0115 \\
\vspace{-1.8mm}
CDM &  0.0110\\
\vspace{-1.8mm}
2nd order CDM & 0.0116\\
\hline
\vspace{-1.8mm}
FDM w/o ${}^b\vv^{t-1}_{\text{ref}}, {}^b\vv^{t-1}_{\text{com}}$ & 0.0406 \\
\vspace{-1.8mm}
FDM w/o $\pv_{\text{foot}}^t$ & 0.0252 \\
\vspace{-1.8mm}
FDM & 0.0199 \\
\vspace{-1.8mm}
2nd order FDM & 0.0249\\
\hline
\vspace{-1.8mm}
CVM w/o ${}^b\vv^{t-1}_{\text{ref}}, {}^b\vv^{t-1}_{\text{com}}$ & 0.118 \\
\vspace{-1.8mm}
CVM w/o $\pv_{\text{foot}}^t$ & 0.0540 \\
\vspace{-1.8mm}
CVM & 0.0492 \\
\vspace{-1.8mm}
2nd order CVM & 0.0610\\
\hline
\end{tabular}
\label{table:cdm_model}
\end{table}

\begin{table*}[t]
\caption{Experimental results of the proposed approach on three different indoor scenarios.}
\centering
\begin{tabular}{ |p{3cm}|p{1cm}|p{1cm}|p{1cm}|p{1cm}|p{1cm}|p{1cm}|p{1cm}|p{1cm}|p{1cm}|  }
\hline
\multicolumn{1}{|c|}{} &
\multicolumn{3}{|c|}{\textbf{Scenario I}} &
\multicolumn{3}{c|}{\textbf{Scenario II}} &
\multicolumn{3}{c|}{\textbf{Scenario III}} \\
\hline
\textbf{Horizon}   & \textbf{3}  & \textbf{5} &  \textbf{7} & \textbf{3}  & \textbf{5} &  \textbf{7} & \textbf{3}  & \textbf{5} &  \textbf{7} \\
\hline
Success (\%)  & 100\% & 100\% & 100\% & 100\% & 100\% & 100\% & 100\% &  100\% & 100\%  \\
\hline
Planning Time (ms) &  0.239   &  0.924 & 1.75 & 0.475 & 0.765 & 0.975 & 0.234 & 0.825 & 1.20\\
\hline
Cumulative Cost & 32.6 & 23.96 & 20.95 & 25.2 & 20.5 & 18.1 & 32.3 & 26.3 & 24.3\\
\hline
CDM MAE (m) &  \multicolumn{3}{c|}{0.00786} & \multicolumn{3}{c|}{0.00360} & \multicolumn{3}{c|}{0.00339} \\
FDM MAE (m) &  \multicolumn{3}{c|}{0.0198} &  \multicolumn{3}{c|}{0.0203} &  \multicolumn{3}{c|}{0.0170}\\
CVM MAE (m/s) & \multicolumn{3}{c|}{0.0317} & \multicolumn{3}{c|}{0.0369} &  \multicolumn{3}{c|}{0.0445} \\
\hline
\end{tabular}
\label{table:exp_results}
\end{table*}

\subsection{Motion Models}

We use the Root Mean Squared Error (RMSE) as a metric for the models' evaluation. In particular, for each model we compare the obtained RMSE for different input choices on the test set. Moreover, we also compare the best obtained linear model with a non-linear one.

For each model, we use as a baseline the input definition given in Section~\ref{sec:approach}, and compare it to three other formulations, which are:

\begin{enumerate}
  \item without considering ${}^b\vv^{t-1}_{\text{ref}}, {}^b\vv^{t-1}_{\text{com}}$
  \item without considering $\pv_{\text{foot}}^t$
  \item 2nd-order polynomial of the state~(\ref{eq:state2})
\end{enumerate}
where 1) and 2) respectively remove ${}^b\vv^{t-1}_{\text{ref}}$, ${}^b\vv^{t-1}_{\text{com}}$ and $\bar{\pv}_{\text{foot}}^t$ from the input definition given in Section~\ref{sec:approach}, and 3) uses both a linear and a quadratic expression of the input space. Table~\ref{table:cdm_model} shows respectively the comparison of the RMSE for the aforementioned formulations.

We note how the trend is similar for all three models. More precisely, we observe that the worst performing linear model is the one that does not consider ${}^b\vv^{t-1}_{\text{ref}}$ and ${}^b\vv^{t-1}_{\text{com}}$. This can be explained by the fact that Aliengo's controller does not track aggressively the reference velocity given at time $t$ and its behavior is influenced by the velocity history at time $t-1$. Similarly, we also note how not considering $\bar{\pv}_{\text{foot}}^t$ decreases the models' accuracy, although not as significantly for the CDM as for the rest of the models. This can be explained by the fact that the controller's behavior and consequently the CoM displacement, feet displacement, and velocity change depends on the postural configuration brought by the feet.

Finally, we also compare the best linear model obtained with a second-order polynomial to see if a non-linear formulation could improve the accuracy of the regressor. For all three models we obtain the same result, where the linear model has a higher accuracy than the non-linear one. This emphasizes the fact that there is a strong linear relationship between the input variables and output variables. Hence, even trying highly non-linear models would not bring any useful benefit to the models.

\subsection{Motion Planning Pipeline}
The motion planning pipeline is evaluated on three different real environments with varying degrees of complexity. Their morphology along with their respective foot costmap can be observed in Fig.~\ref{figure:scenarios}.

The first scenario (Scenario I) consists of two symmetrical gaps across the longitudinal axis of the robot, respectively of $10$~cm and $15$~cm, while the second scenario (Scenario II) contains two asymmetrical gaps of $15$~cm and a small step of $6$~cm. These scenarios are proposed in order to highlight the particular capability of the built-in controller. As explained in Sect.~\ref{sec:formulation}, the robot can only perform a trotting gait, and thus the swing feet are constrained to achieve the same foot swing distance during normal operation hindering the capability of the robot to traverse asymmetric scenarios. Finally, the third and last experiment (Scenario III) comprises one symmetrical gap of $10$~cm, two asymmetrical gaps of $12$~cm and $13$~cm, and two small steps of $6$~cm. 

The obtained results, which can be observed in Table~\ref{table:exp_results}, are collected during three different attempts for each scenario. The robot is able to successfully complete all the proposed scenarios, and as expected, varying the length of the planning horizon has a direct impact of the optimality of the final solution. For this, we report in Table~\ref{table:exp_results} the cumulative cost obtained considering only the first step performed by the robot at the end of each re-planning procedure. Furthermore, we also observe how the planner takes on average less than $2$ms for the re-planning phase even with longer horizons, making it suitable for reactive motions.
Finally, we analyze the Mean Absolute Errors (MAE) of the three learned models on the three proposed scenarios. Even if their values are small, the data demonstrates that re-planning is necessary in order to avoid the inevitable error propagation.  

Videos of the experiments and comparisons with the blind built-in controller are included in the accompanying video \footnote{\url{https://tinyurl.com/4rwsthr7}}. As expected, without visual information, the robot is more statistically prone to failure in such scenarios. Furthermore, we include in the accompanying video an additional experiment in a stair-like scenario to show the potentiality of the proposed method.

\section{Conclusions}
In this paper, we proposed a vision-based planning method able to enhance the built-in blind controller of Aliengo. To this end, the black-box nature of the controller is learned by means of three different regressors, fully integrated into the motion planning pipeline. Finally, we conducted experiments on scenarios with different complexity to highlight the capability of the approach.

The limitations of the proposed approach lie on the limited authority that the user can have on the black-box controller. In fact, to traverse some scenarios, it could be needed not only to modify the footstep location but even the step height (e.g. for high stairs) or the trunk pose~\cite{fahmi2022vital} in advance. Both of these possibilities, in the case of Aliengo, are limited or precluded for the control modality used in this approach.

Future works will focus on reducing the conservativeness of the method. The safety margin (Sect.~\ref{sec:approach}) introduced 
 for coping with prediction errors can be modified during planning, for example, by applying Bayesian regression techniques~\cite{deisenroth_MLBook} and actively considering the additional information on the regressor uncertainty. Furthermore, we plan to enhance the ease of use of the proposed approach. The goal position could be in fact be computed directly by joystick commands, giving the possibility to the user to control the robot in the easiest and most reactive way possible.    
\label{sec:conclusions}

\addtolength{\textheight}{-12cm}   % This command serves to balance the column lengths
                                  % on the last page of the document manually. It shortens
                                  % the textheight of the last page by a suitable amount.
                                  % This command does not take effect until the next page
                                  % so it should come on the page before the last. Make
                                  % sure that you do not shorten the textheight too much.

%%%%%%%%%%%%%%%%%%%%%%%%%%%%%%%%%%%%%%%%%%%%%%%%%%%%%%%%%%%%%%%%%%%%%%%%%%%%%%%%

%%%%%%%%%%%%%%%%%%%%%%%%%%%%%%%%%%%%%%%%%%%%%%%%%%%%%%%%%%%%%%%%%%%%%%%%%%%%%%%%

\bibliographystyle{IEEEtran}
\bibliography{bibliography}

\end{document}

%% file: boldfacemath.tex
\usepackage{amsmath}

\newcommand{\vect}[1]{\boldsymbol{#1}}
\newcommand{\mat}[1]{\boldsymbol{#1}}

\newcommand{\diffs}[3]{\frac{\partial^2 #1}{
\ifx#2#3 
\partial #2^2
\else
\partial #2 \partial #3
\fi
}}
\newcommand{\fv}{\vect{f}}

\newcommand{\pv}{\vect{p}}

\newcommand{\sv}{\vect{s}}

\newcommand{\vv}{\vect{v}}

\newcommand{\xv}{\vect{x}}

\newcommand{\deltav}{\vect{\delta}}

\newcommand{\Rm}{\mat{R}}